\documentclass[sigconf]{acmart}
\usepackage{booktabs} 
\usepackage{bm}
\usepackage{url}
\usepackage{mathrsfs}
\usepackage{multirow}
\usepackage{balance}
\usepackage{amsthm}
\usepackage{amsmath}
\usepackage{amstext}
\usepackage{amssymb}
\usepackage[ruled,vlined,linesnumbered]{algorithm2e}
\usepackage{subcaption}
\usepackage{overpic}
\usepackage{color}
\usepackage{array}
\usepackage{enumitem}
\newcommand{\PreserveBackslash}[1]{\let\temp=\\#1\let\\=\temp}
\newcolumntype{C}[1]{>{\PreserveBackslash\centering}p{#1}}
\newcolumntype{R}[1]{>{\PreserveBackslash\raggedleft}p{#1}}
\newcolumntype{L}[1]{>{\PreserveBackslash\raggedright}p{#1}}

\copyrightyear{2019}
\acmYear{2019} 
\setcopyright{iw3c2w3}
\acmConference[WWW '19]{Proceedings of the 2019 World Wide Web Conference}{May 13--17, 2019}{San Francisco, CA, USA}
\acmBooktitle{Proceedings of the 2019 World Wide Web Conference (WWW '19), May 13--17, 2019, San Francisco, CA, USA}
\acmPrice{}
\acmDOI{10.1145/3308558.3313545}
\acmISBN{978-1-4503-6674-8/19/05}

\usepackage{balance}

\begin{document}
\title{On Attribution of Recurrent Neural Network Predictions \\ via Additive Decomposition}

\author{Mengnan Du, Ninghao Liu, Fan Yang, Shuiwang Ji, Xia Hu}
\affiliation{
 \institution{Department of Computer Science and Engineering, Texas A\&M University} 
 \city{} 
 }
\email{{dumengnan, nhliu43, nacoyang, sji, xiahu}@tamu.edu}

\renewcommand{\shortauthors}{MN. Du et al.}

\begin{abstract}
RNN models have achieved the state-of-the-art performance in a wide range of text mining tasks. However, these models are often regarded as black-boxes and are criticized due to the lack of interpretability. In this paper, we enhance the interpretability of RNNs by providing interpretable rationales for RNN predictions. Nevertheless, interpreting RNNs is a challenging problem. Firstly, unlike existing methods that rely on local approximation, we aim to provide rationales that are more faithful to the decision making process of RNN models. Secondly, a flexible interpretation method should be able to assign contribution scores to text segments of varying lengths, instead of only to individual words. To tackle these challenges, we propose a novel attribution method, called REAT, to provide interpretations to RNN predictions. REAT decomposes the final prediction of a RNN into additive contribution of each word in the input text. This additive decomposition enables REAT to further obtain phrase-level attribution scores. In addition, REAT is generally applicable to various RNN architectures, including GRU, LSTM and their bidirectional versions. Experimental results demonstrate the faithfulness and interpretability of the proposed attribution method. Comprehensive analysis shows that our attribution method could unveil the useful linguistic knowledge captured by RNNs. Some analysis further demonstrates our method could be utilized as a debugging tool to examine the vulnerability and failure reasons of RNNs, which may lead to several promising future directions to promote generalization ability of RNNs.
\end{abstract}

%
%



\maketitle

\section{Introduction}
The RNNs, such as LSTM~\cite{hochreiter1997long} and GRU~\cite{cho2014learning}, have achieved the state-of-the-art performance in a variety of text classification tasks. These deep text classification models have became increasingly deployed in different web applications, including sentiment classification~\cite{tang2015document}, named entities recognition~\cite{peters2017semi}, textual entailment~\cite{chen2016enhanced}, etc. Despite the superior performance, RNNs are often criticized by their lack of interpretability, and are often treated as black-boxes~\cite{lipton2016mythos}. The lack of understanding of the mechanism behind RNN predictions not only reduces the acceptance from end-users towards the deployed predictors, but also limits ability of system developers in diagnosing the model, searching the reasons for wrong predictions, and improving the model architectures. Therefore, it is highly desirable to explore the interpretability of RNNs, so as to provide insight of how they process text inputs and make inferences therefrom.

It is possible to provide interpretability for RNNs from two perspectives: adding interpretable components to RNN models or performing post-hoc attribution~\cite{du2018techniques}. The former category integrates interpretability directly into the structure of deep models, and usually resorts to attention mechanism~\cite{lin2017structured,yang2016hierarchical}. However, the attention components only provide an indirect indicator of contribution scores. For classification tasks, it is not clear which specific class these attention weights contribute to~\cite{murdoch2018beyond}. In contrast, the philosophy of post-hoc attribution is to provide interpretation to predictions made by a pre-trained black-box model without any architectural modification. Post-hoc attribution aims to attribute the prediction of a model to its input features, e.g., words in a text classification task, and produce a heatmap as interpretation, indicating the contribution of each feature to a particular class of interest. In this paper, we follow the post-hoc attribution strategy, as it could provide understandable rationale to the given predicted label from the model while keeping the underlying model status intact.

Although post-hoc attribution has been extensively studied for understanding MLP and CNNs~\cite{du2018towards}, attribution for RNN predictions is still a technically challenging problem. First, one challenge lies in how to guarantee that the interpretations are indeed faithful to the original model. Many previous attribution work, including back-propagation based methods~\cite{hechtlinger2016interpretation,denil2014extraction}, perturbation based methods~\cite{li2016understanding,kadar2017representation}, and local approximation based methods~\cite{ribeiro2016should,ribeiro2018anchors}, all follow the philosophy of \emph{local interpretation}. That is, they generate interpretable approximation of the original model around the neighborhood of a given prediction to be explained. However, it is not guaranteed that the generated interpretations accurately reflect the decision making process of the original model~\cite{chu2018exact,guo2018lemna}. In this case, it is hard to tell whether an unexpected interpretation is caused by misbehavior of the model or limitation of the attribution method. Second, it is challenging to develop a flexible attribution method which could generate attribution scores to text segments (e.g., phrases) of varying lengths. Prior work for RNNs attribution mainly focuses on identifying \emph{word-level contribution scores}~\cite{denil2014extraction,sundararajan2017axiomatic,ribeiro2016should}, which assigns a real-value score for each of the words, indicating the extent to which it contributes to a particular prediction. However, word-level attributions fail to explain why RNNs are successful to process sequences where the order of the data entries matters. Consider an example of sentence sentiment analysis, ``I do not dislike cabin cruisers.'' expresses a neutral opinion. Word-level attribution methods may accurately identify that the word ``dislike'' has a negative contribution for this prediction, but they fail to capture that the negation word ``not'' has shifted its polarity and the word combinations ``not dislike'' have a positive impact for model prediction.

In this paper, we propose a decomposition based attribution method, called REAT (\underline{RE}current \underline{AT}tribution Method), to provide interpretation for given predictions made by a RNN in a faithful and flexible manner. Through modeling the information flowing process of the hidden state vectors in RNN models, REAT could decompose the final prediction of a RNN into additive contribution of each word in the input text. Since REAT is constructed by directly leveraging the information propagation process from hidden state vectors to the output layer, it enjoys the benefit of high faithfulness to the original RNN model. This method not only can quantify the contribution of each individual word to a prediction, but also could naturally be applied to identify the contribution of word sequences. It thus enables the illumination of how RNNs make use of sequential information, as well as how they capture long-term dependencies. In addition, REAT is widely applicable to different recurrent architectures, including LSTMs and GRUs, and their bidirectional versions. Furthermore, based on the observation that language usually exhibits hierarchical structures, we expand REAT to a hierarchical attribution method, to represent the contributions at different levels of granularity. 
The major contributions of this paper are summarized as follows: 
\begin{itemize}[leftmargin=*]
\item We propose a RNN attribution method based on additive decomposition, called REAT, which could provide both word-level and phrase-level attribution scores.
\item REAT is applicable to different RNN architectures. We demonstrate its applicability to three standard RNN architectures, including GRU, LSTM and Bidirectional GRU.
\item Experimental results on two sentiment analysis datasets validate the faithfulness and interpretability of the proposed method.
\item We demonstrate that REAT could be utilized as a debugging tool to analyze the useful linguistic knowledge captured by RNNs and examine the vulnerability and failure reasons of RNNs.
\end{itemize}

\section{Preliminaries}
In this section, we start by introducing post-hoc attribution that serves as the basic attribution scheme in this paper. Then we introduce three representative RNN architectures as well as the output layer which transforms the hidden state into probability output.

\subsection{Post-hoc Attribution}
\noindent\textbf{Notations}: Consider a typical multi-class text classification task, a RNN-based classification model can be denoted as $f : X \rightarrow Y$, where $X$ is the text space, and $Y = \{1,...,C\}$ denotes the set of output classes. The RNN model accepts an instance $ \textbf{x} \in X$ as input, and maps it to an output class: $f(\textbf{x})= c \in Y$. Assume the input is composed of a sequence of $T$ words: $\textbf{x} = \{x_1, ..., x_T \}$ and each word $x_t \in \mathbb{R}^d$ denotes the embedding representation of the $t$-th word. The high level idea of post-hoc attribution is to attribute the prediction $f(\textbf{x})$ of a RNN model to its input features $\textbf{x}$ and output a heatmap indicating the contribution of each feature $x_t \in \textbf{x}$ to a particular class of interest $c$. Specifically, we target to generate attributions for a RNN prediction which is specified as follows:

\noindent\textbf{Phrase-level Attribution}: We first partition the input text into meaningful phrases (coherent pieces of text) and then attach an attribution score to each individual phrase. Given one index phrase as query. The index set of a  phrase that we want to calculate its attribution score is denoted as: $A = \{q,...r \}$ where $1 \leq q \leq r \leq T$. We indicate the attribution score for the targeting phrase as $S(\textbf{x}_A)$.

\subsection{RNN Architectures}\label{architectures}
RNNs come in many variants with different architectures, which results in different mapping functions $f$. In this paper, we discuss three representative RNN architectures that are fundamental and have been widely used in many applications. The common formulations in different RNN architectures motivate the design of our interpretation approach.

\noindent\textbf{LSTM}:
In a Vanilla RNN, at any time step in a sequence, the hidden state $h_t$ is calculated based on its previous hidden state $h_{t-1}$ and the current input vector $x_t$:
\begin{equation}
h_t = \text{tanh}(W_{hx}x_t + W_{hh}h_{t-1} + b_{h}).
\label{equ:RNNupdate}
\end{equation}
Comparing to the Vanilla RNN, LSTM makes two major changes. Firstly, LSTM introduces a cell state $c_t$ that serves as an explicit memory. Secondly, instead of simply updating the hidden state $h_t$ in Eq. (\ref{equ:RNNupdate}), LSTM uses three gates: input gate $i_t$, forget gate $f_t$, and output gate $o_t$ to update the hidden state $h_t$.
\begin{equation}
\begin{aligned}
\centering
& i_t = \sigma(W_{ix} x_t + W_{ih} h_{t-1} + b_i ) \\
& f_t = \sigma(W_{fx} x_t + W_{fh} h_{t-1} + b_f ) \\
& o_t = \sigma(W_{ox} x_t + W_{oh} h_{t-1} + b_o ) \\
& g_t = \text{tanh}(W_{gx} x_t + W_{gh} h_{t-1} + b_g ) \\
& c_t = f_t \odot c_{t-1} + i_t \odot g_t \\
& h_t = o_t \odot \text{tanh}(c_t).
\end{aligned}
\label{equ:LSTMupdate}
\end{equation}
where each $W$ and $b$ represent weight matrix and bias vector respectively, and $\odot$ denotes element-wise multiplication.

\noindent\textbf{GRU}: GRU makes some slight modifications on the basis of LSTM. It only has two gates, i.e., reset gate $r_t$ and update gate $u_t$. Besides, it merges the cell state and hidden state into a single hidden state $h_t$. The updating rule of hidden state is denoted as follows.
\begin{equation}
\begin{aligned}
\centering
& r_t = \sigma(W_{rx} x_t + W_{rh} h_{t-1} + b_r ) \\
& u_t = \sigma(W_{ux} x_t + W_{uh} h_{t-1} + b_u ) \\
& g_t = \text{tanh}(W_{gx} x_t + r_t \odot W_{gh} h_{t-1} + b_g ) \\
& h_t =  u_t \odot h_{t-1} +  (1-u_t) \odot g_t.
\end{aligned}
\label{equ:GRUupdate}
\end{equation}

\noindent\textbf{Bidirectional GRU}: This model is constructed by putting two independent GRUs together. The word sequences are fed into one GRU in normal time order, and in reverse time order into another. For each network, the hidden state is updated using the same rule as Eq. (\ref{equ:GRUupdate}). For the sake of brevity, we use subscripts $n$ and $r$ to represent the $normal$ and $reverse$ network respectively. For classification tasks, the final hidden vector fed into the output layer is constructed by concatenating the hidden vector at time step $T$ for the normal GRU and the hidden vector at time step $1$ for the reverse one:
\begin{equation}
h = h_{T,n} \oplus h_{1,r},
\end{equation}
where symbol $\oplus$ denotes concatenation operation of two vectors. Both $h_{T,n}$ and $h_{1,r}$ contain the information of the whole text. Hidden state $h_{T,n}$ encodes the information of the normal text (from $x_1$ to $x_T$), while $h_{1,r}$ captures the information in the inverse text (from $x_T$ to $x_1$). In the remaining part of this paper, we use BiGRU to denote Bidirectional GRU.

\subsection{RNN Output Layer}
To serve the purpose of multi-class text classification, a discriminative layer is added after the activation vector $h_T$ of the last hidden layer at time step $T$. This layer takes the hidden state $h_T$ as input and turns it into a logit vector $z$ using a weight matrix $W$:
\begin{equation}
z = W h_T,
\label{equ:logits}
\end{equation}
and then produces the class probabilities using a softmax layer which converts the logit $z_c$ for a class $c$ into a probability $y_c$, by comparing it with other logit values:
\begin{equation}
y_c = \text{softmax}(z)_c = \frac{\text{exp}(z_c)}{\sum_{k=1}^C \text{exp}( z_k )}.
\end{equation}

\section{Methodology}
In this section, we introduce the proposed attribution method for explaining RNN predictions. We first present a general method for phrase-level RNN prediction attribution, and then apply it to three widely used RNN architectures. Finally, we expand the method to enable hierarchical attribution schemes.

\subsection{Phrase-level Attribution Method for Recurrent Models}

\subsubsection{A Naive Attribution Approach} \label{sec:naiveMethod}
From the last section, we know that RNNs possess a series of hidden state vectors $\{h_t\}_{t=1,...,T}$, where each vector $h_t$ stores information about the past input blocks, ranging from time step 1 to $t$. A crucial property of the hidden state vector is that it is updated from time step to time step. Knowing how much information is accumulated at each time step enables us to derive the contribution of that time step towards the final prediction. Intuitively, we denote the response of the RNN model to word $\textbf{x}_t$, and thereby the information gained at step $t$, as follows:
\begin{equation}
\tilde{h}_t = g(\textbf{x}_t) =h_t - h_{t-1}.
\label{equ:nativeupdate}
\end{equation}

In this way, we can consider the final hidden state vector $h_T$ to contain information accumulated at $T$ time steps, denoted as: $h_T = \sum_{t=1}^T \tilde{h}_t = \sum_{t=1}^T (h_t - h_{t-1})$. Then, we can decompose the logit $z_c$ (see Eq. (\ref{equ:logits}) ) for target class $c$ using a sequence of factors:
\begin{equation}
z_c = W_c h_T = \sum_{t=1}^T W_c(h_t - h_{t-1}).
\end{equation}
It can be treated as the additive contribution of each word in the input $\textbf{x}$ to the output logit (unnormalized output probability) of class $c$. Therefore, the contribution of word $\textbf{x}_t$ towards the logit $z_c$ can be calculated as: 
\begin{equation}
S (\textbf{x}_t) = W_c(h_t - h_{t-1}).
\label{equ:baseline}
\end{equation}
However, this decomposition has a severe shortcoming. The underlying assumption of this formulation is that 
all evidence accumulated up to time step $t-1$ has been transferred to time $t$. This violates the updating rules of all the three architectures listed in Sec.~\ref{architectures}. Actually, it fails to take into consideration the forgetting mechanism of RNNs.
For example, both LSTM and GRU have explicit gates to model the forgetting and remembering mechanism, which serves the purpose of controlling information flow and calculating how much proportion of information derived from previous time steps should be kept.

\begin{figure}
  \centering
  \includegraphics[width=0.55\linewidth]{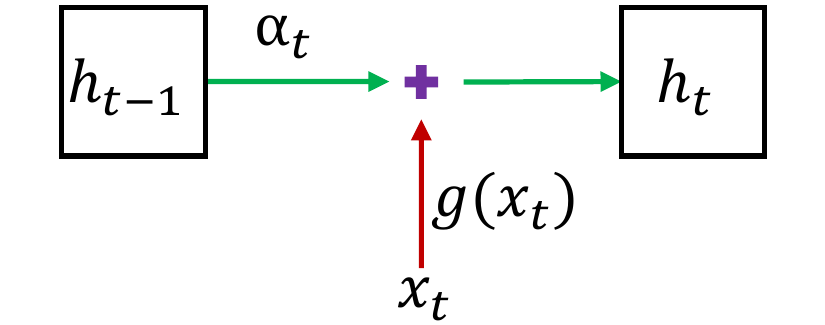}
  \caption{An illustration of the summarized updating rule for the RNN hidden state vectors. }
  \label{fig:updating}
\end{figure}
\subsubsection{The Proposed Recurrent Attribution Method (REAT)}
Many variants of RNNs share a similar form of hidden state updates. We summarize a common rule which can be applied to different recurrent network architectures, as illustrated in Fig.~\ref{fig:updating}. The rule maintains a hidden state $h_t$ which summarizes information for past sequences. Also, at each time step $t$, the hidden state is updated using the following equation:
\begin{equation}
h_t = \alpha_t \odot h_{t-1} + \tilde{h}_t,
\label{equ:hiddenupdate}
\end{equation}
where $\alpha_t \in [0,1]^{d'}$ so that only partial evidence obtained by RNN from previous $t-1$ steps is brought to the time step $t$. Here $d'$ is the dimension of hidden state vectors. A higher value of $\alpha$ means that the RNN model preserves more important information from previous time steps. 
Here $\tilde{h}_t = g(\textbf{x}_t)$ denotes the evidence that a RNN obtains at time step $t$, but note that it is no longer defined as in Eq. (\ref{equ:nativeupdate}). Some RNN architectures obey this rule exactly, like GRU, while some other architectures follow this rule approximately, such as LSTM.
Based on this hidden state updating rule, we can iteratively trace back the generation of $h_T$ and decompose the logit value $z_c$ into the following form:
\begin{equation}
 z_c = W_c h_T  =   \sum_{t=1}^T W_c (\tilde{h}_t \odot \prod_{k=t+1}^T \alpha_k).
\label{equ:decomposition1}
\end{equation}
Replacing each $\tilde{h}_t$ with $h_t - \alpha_t \odot h_{t-1}$, we can reformulate the additive decomposition as:
\begin{equation}
 z_c =  \sum_{t=1}^T W_c [(h_t - \alpha_t \odot h_{t-1})\odot \prod_{k=t+1}^T \alpha_k].
\label{equ:decomposition2}
\end{equation}
The main benefit of Eq.~(\ref{equ:decomposition2}) comparing to Eq.~(\ref{equ:decomposition1}) is that we do not need to know the exact form of $\tilde{h}_t$. Merely knowing the hidden state vector $h_t$ and the updating parameter vector $\alpha_t$ will be sufficient to derive the decomposition. Eq.~(\ref{equ:decomposition2}) can be considered as the additive contribution of each word $\textbf{x}_t$ towards the logit $z_c$, which is the unnormalized probability for target class $c$. By taking out the term relevant to time step $t$, we can derive the contribution value for a single word $\textbf{x}_t$ at the current time step:
\begin{equation}
 S (\textbf{x}_t)= W_c [\underbrace{ (h_t - \alpha_t \odot h_{t-1})}_{Updating} \odot \underbrace{\prod_{k=t+1}^T \alpha_k }_{Forgetting}].
 \label{equ:wordattribution}
\end{equation}
The above formulation within the square brackets is the element-wise multiplication of two terms. The left term denotes the updating evidence from time $t-1$ to $t$, i.e., the contribution to class $c$ by the input word $\textbf{x}_t$. The right term represents the forgetting mechanism of RNN.
The evidence that a RNN has gathered at time step $t$ gradually diminishes as the time increases from $t+1$ to the final time step $T$. That is to say, only part of the updating evidence will have impacts on the classification task at time step $T$.

\begin{figure}
  \centering
  \includegraphics[width=1.0\linewidth]{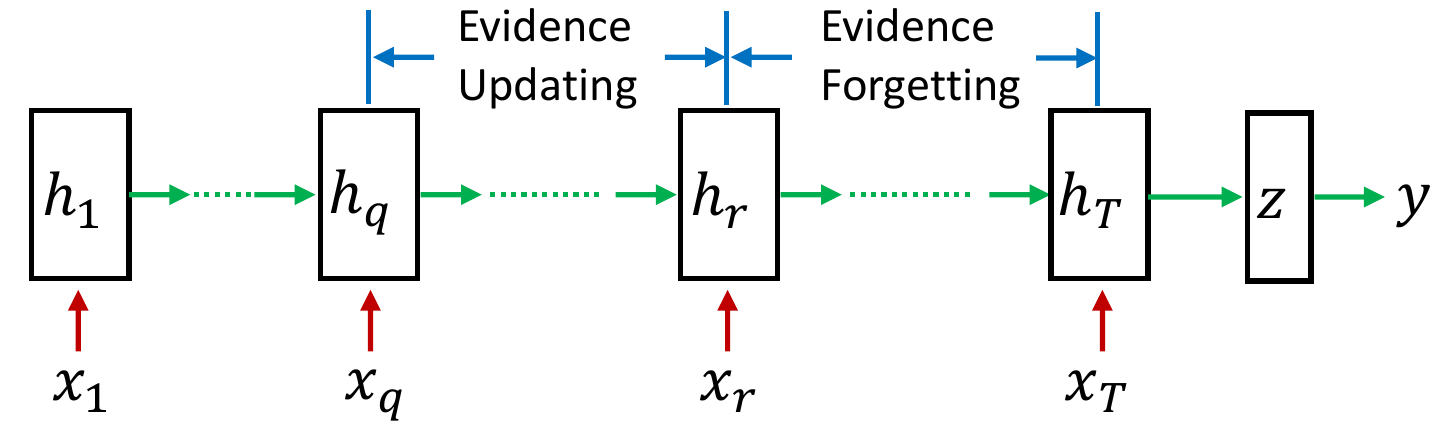}
  \caption{An illustration of the proposed method. }
  \label{fig:pipeline}
\end{figure}
Based on the word-level additive attribution formulation in Eq.~(\ref{equ:wordattribution}), we can conveniently derive the phrase-level attribution.
For a phrase $\textbf{x}_A$, where $A = \{q,...r \}$, $1 \leq q \leq r \leq T$, its attribution score $S(\textbf{x}_A)$ can be denoted as:
\begin{equation}
 S (\textbf{x}_A)=  W_c [\underbrace{(h_r - \prod_{j=q}^r \alpha_j \odot h_{q-1})}_{Updating} \odot \underbrace{\prod_{k=r+1}^T \alpha_k}_{Forgetting}].
\label{equ:phraseattribution}
\end{equation}
Similar to word-level attribution, phrase-level attribution $S(\textbf{x}_A)$ also contains two terms. The left term within the square brackets in Eq.~(\ref{equ:phraseattribution}) represents the updating evidence from time step $q-1$ to time $r$, while the right term denotes how much percentage of the evidence has been forgotten from time $r+1$ to $T$. We illustrate this process in Fig.~\ref{fig:pipeline}.
It is worth noting that the key component here to derive phrase-level attribution score for a RNN classifier is to obtain the hidden state vectors and the updating vectors. Usually, only one feed forward operation is needed to derive phrase-level attribution score, which can be implemented efficiently.

\subsection{Applications to Specific Architectures}
In this section, we apply the proposed idea to three standard RNN architectures, including LSTM, GRU and BiGRU.

\subsubsection{GRU Attribution}\label{sec:GRUattribution}

The hidden state vector updating rule for GRU is written as:
\begin{equation}
 h_t = u_t \odot h_{t-1} +  (1-u_t) \odot g_t,
\end{equation}
which conforms with the paradigm in Eq. (\ref{equ:hiddenupdate}). Therefore, for a phrase $\textbf{x}_A$, where $A = \{q,...r \}$, $1 \leq q \leq r \leq T$, we can directly replace $\alpha_t$ in Eq. (\ref{equ:phraseattribution}) with the updating gate vector $u_t$ of the GRU model, and obtain the phrase-level attribution score $S(\textbf{x}_A)$:
\begin{equation}
 S (\textbf{x}_A)= W_c [(h_r - \prod_{j=q}^r u_j \odot h_{q-1}) \odot \prod_{k=r+1}^T u_k].
\end{equation}

\subsubsection{LSTM Attribution}
Although it is difficult to directly match the LSTM updating rule of hidden state $h_t$ in Eq. (\ref{equ:LSTMupdate}) to the paradigm in Eq.~(\ref{equ:hiddenupdate}),
the update of the cell state $c_t$ adheres to the expected updating format in Eq.~(\ref{equ:hiddenupdate}). It is denoted as: $c_t = f_t \odot c_{t-1} + i_t \odot g_t$, where $f_t$ is the forgetting gate of LSTM.

Based on the updating rule from the cell state $c_t$ to hidden state: $h_t = o_t\odot tanh(c_t)$, approximately we can obtain:
\begin{equation}
 c_t  \sim  \frac{h_t}{o_t},
\end{equation}
where the right term is a element-wise division. Thus we approximate the updating rule of the hidden state vector for LSTM as below:
\begin{equation}
 h_t = \frac{f_t \odot o_t}{o_{t-1}} \odot h_{t-1} + \tilde{h}_t.
\end{equation}
Then, we can further decompose the final hidden state vector $h_T$ into the following formulation:
\begin{equation}
 h_T  =  \sum_{t=1}^T (h_t - \frac{f_t \odot o_t}{o_{t-1}} \odot h_{t-1})) \odot \prod_{k=t+1}^T \frac{f_k \odot o_k}{o_{k-1}}.
\end{equation}
For a phrase $\textbf{x}_A$, where $A = \{q,...r \}$, $1 \leq q \leq r \leq T$, we can obtain the attribution score $S(\textbf{x}_A)$:
\begin{equation}
 S (\textbf{x}_A)= W_c [(h_r - \prod_{j=q}^r \frac{f_j \odot o_j}{o_{j-1}} \odot h_{q-1}) \odot \prod_{k=r+1}^T \frac{f_k \odot o_k}{o_{k-1}}].
\end{equation}

\subsubsection{BiGRU Attribution}
In this section, we illustrate how the proposed method can be applied to bidirectional architectures.
The last hidden state $h_{T,n}$ at time step $T$ of the normal GRU uses the identical decomposition as in Sec. \ref{sec:GRUattribution}. As for the reverse GRU, we use the hidden state vector at time step $1$ in order to capture the information from $1$ to $T$, which can be decomposed as follows:
\begin{equation}
 h_{1,r}  =  \sum_{t=1}^T   (h_{t,r} - u_{t,r} \odot h_{t+1,r}) \odot \prod_{k=1}^{t-1} u_{k,r}.
\label{equ:cellDecompositionReverse}
\end{equation}

Recall that the final hidden vector fed into the classification layer is the concatenation of the hidden vector at time step $T$ for the normal GRU and the hidden vector at time step $1$ for the reverse GRU, i.e., $h = h_{T,n}\oplus h_{1,r}$. As such, the logit value is computed as $z_c = W_c(h_{T,n}\oplus h_{1,r})$.
To obtain the attribution score for a phrase $\textbf{x}_A$, where $A = \{q,...r \}$, we first calculate the updated hidden evidence for the normal network and the reverse network respectively. We then concatenate these two updated hidden evidence, and multiply it with $W_c$ to produce the contribution scores.
\begin{equation}
\begin{aligned}
\centering
 S (\textbf{x}_A)= &   W_c [((h_{r,n} - \prod_{j=q}^r u_{j,n} \odot h_{q-1,n}) \odot \prod_{k=r+1}^T u_{k,n})\oplus \\
 & ((h_{q,r} - \prod_{j=q}^r u_{j,r} \odot h_{r+1,r}) \odot \prod_{k=1}^{q-1} u_{k,r} )].
\end{aligned}
\label{equ:bilstmAttribution}
\end{equation}

\begin{figure}[t]
  \centering
  \includegraphics[width=0.8\linewidth]{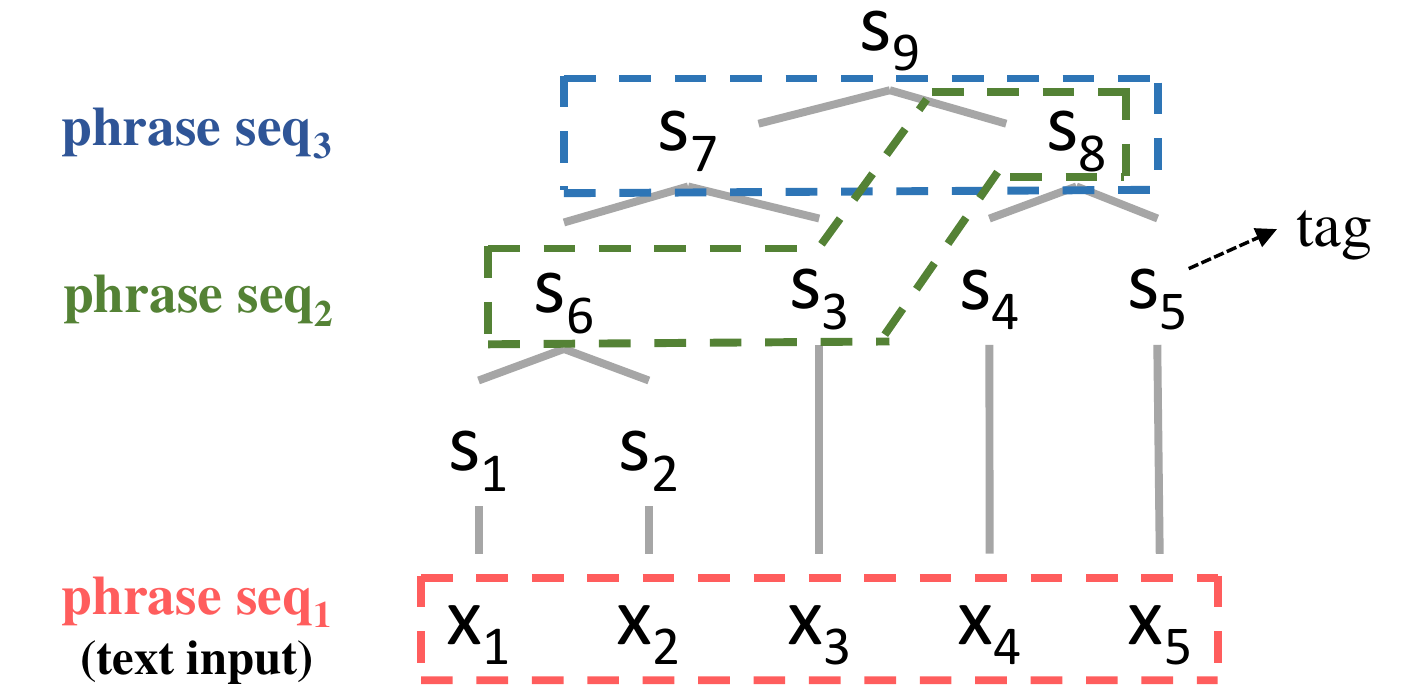}
  \caption{The proposed phrase-based attribution method is naturally compatible with sentence parsing. }
  \label{fig:parsingTree}
\end{figure}

\begin{table}
\small
\centering
\begin{tabular}{L{2.3cm}L{5.5cm}}
\toprule
&\textbf{Regular expression}\\
\midrule
\textbf{ {Verb chunks}}& $<VERB>*<ADV>*<PART>*<VERB>+<PART>*$ \\
\textbf{  {Noun chunks}}& $<DET>? (<NOUN>+ <ADP|CONJ>)* <NOUN>+$\\
\bottomrule
\end{tabular}
\caption{Regular expression matching.}
\vspace{-10pt}
\label{tab:regularexpression}
\end{table}

\subsection{Automatic Attribution Generation}
The developed method so far can only calculate the contribution score of a phrase whose beginning and ending in $\textbf{x}$ are already specified. A natural question that arises is: how to comprehensively analyze and combine the contributions of different parts of an input text, so as to explain the response of RNNs spanning on the whole text. A naive way is to exhaustively select phrases of different length and starting points, and compute their contribution scores respectively. However, such search scheme will span over a lot of redundant and trivially overlapping text segments. In this case, the redundant amount of interpretation information may be overwhelming to users. Traditional word-level interpretation can be seen as a special case of our method by simply setting the phrase length as one, but its flexibility is limited.

To design an interpretation scheme with both desirable understandability and flexibility, we apply text parsing into the attribution method. The general idea is illustrated in Fig.~\ref{fig:parsingTree}. The parse tree is derived from certain existing text parsing algorithm. Here $\textbf{x}_n$ denotes a word, and $\textbf{s}_i$ denotes a non-terminal symbol (i.e., a POS tag or a phrase type). Our attribution method will be applied to each of the selected phrase sequences. The phrases within each sequence are semantically meaningful and mutually non-overlapping. In this way, we obtain interpretation of different granularity on a sentence. The interpretation is thus more compatible with the cognition habits of human.

In this work, we partition sentences using regular expression matching~\cite{mitkov2005oxford} to divide each sentence into meaningful phrases. Specifically, we use two regular expressions shown in Tab.~\ref{tab:regularexpression} to partition the sentence into continuous noun and verb chunks. Take the sentence ``The movie does n't serve up lot of laughs.'' for example. A middle-level phrase sequence obtained from parsing could be ``\{The movie\} \{does n't serve up\} \{lot of laughs\}.'', which correspond to noun chunk, verb chunk, and noun chunk respectively. Then we feed the sentence into the RNN and the attribution method.
The output is a sequence of phrases along with corresponding attribution scores, capturing the most significant phrases contributing to the network prediction during the prediction lifetime. In practice, each original text may end up with 2 (word-level and phrase-level) or 3 (word-level, phrase-level, and clause-level, where clause denotes part of a sentence or a complete sentence containing a verb) hierarchies, and users can obtain a hierarchical interpretation of the prediction.

\section{Experiments}
In this section, we present the experimental results in order to answer the following research questions (\textbf{RQ}s).

\begin{itemize}[leftmargin=*]
\item \textbf{RQ1} - Does the generated attribution faithfully reflect how the original RNN model makes predictions?
\item \textbf{RQ2} - Does the proposed RNN attribution method outperform alternative approaches in terms of interpretability?
\item \textbf{RQ3} - Could hierarchical interpretation comprehensively capture the evidences for prediction at different levels of granularity?
\item \textbf{RQ4} -  How to use the proposed attribution method to analyze the useful insights captured by RNNs?
\item \textbf{RQ5} - How to use the proposed attribution method to analyze the misbehavior of RNNs and ultimately promote RNNs' generalization capability?
\end{itemize}

\subsection{Datasets}

\begin{table}
\centering
\begin{tabular}{R{3.0cm}C{1.1cm}C{1.1cm}}
\toprule
&\textbf{SST2}&\textbf{Yelp}\\
\midrule
\textbf{\footnotesize{$\#$} \normalsize {training texts}}&6920 & 80000\\
\textbf{\footnotesize{$\#$} \normalsize {development texts}}&872 &1999\\
\textbf{\footnotesize{$\#$} \normalsize {test texts}}&1821 &5492\\
\bottomrule
\end{tabular}
\caption{Dataset Statistics}
\label{fig:datasetStat}
\vspace{-20pt}
\end{table}

We conduct our experiments on two publicly available sentiment analysis datasets, whose statistics are shown in Tab.~\ref{fig:datasetStat}.

\noindent\textbf{Stanford Sentiment Treebank 2 (SST2)}~\cite{socher2013recursive} -
It contains 2 classes (negative and positive). The numbers of instances for training set, development set and test set are 6920, 872, and 1821 respectively.

\noindent\textbf{Yelp Polarity (Yelp)}~\cite{zhang2015character} -
It consists of reviews originally extracted from the Yelp Dataset Challenge 2015 data. Zhang et al.~\cite{zhang2015character} constructed the Yelp reviews polarity dataset by considering stars 1 and 2 as negative, and considering 4 and 5 as positive. The numbers of instances for training set and test set are 560,000 and 38,000 respectively. We use a subset of this dataset by filtering out texts whose length is larger than 40, and then randomly select part of the samples from the training set as development set. Ultimately, the dataset contains 80000, 1999, and 5492 instances for training, development, and test set respectively, with each polarity occupying around half of the instances in each set.

\begin{table}
\centering
\begin{tabular}{R{1.5cm}C{1.5cm} C{1.5cm}}
\toprule
&\textbf{SST2}&\textbf{Yelp}\\
\midrule
\textbf{ {LSTM}}&80.4\%&92.2\%\\
\textbf{  {GRU}}&81.9\% & 90.2\%\\
\textbf{ {BiGRU}}&80.9\%&93.5\%\\
\bottomrule
\end{tabular}
\caption{Prediction accuracy of the three RNN architectures on SST2 and Yelp dataset.}
\vspace{-10pt}
\label{fig:predictionStat}
\end{table}
\subsection{Experimental Setup}
For each classification model used in our experiments, it contains a word embedding layer to transform words to fixed length representation vectors, a recurrent network layer to transform word embeddings to hidden state vectors, and a classification layer for output. Specifically, the pre-trained 300-dimensional word2vec word embedding~\cite{mitkov2005oxford} is utilized to initialize the embedding layer. For those words that do not exist in word2vec, we initialize their embedding vectors with some random values. The dimension of hidden state vectors is 150 for both LSTM and GRU, and 300 for BiGRU. The classification layer is composed of a dense layer and a softmax nonlinear transformation. The Adam optimizer~\cite{kingma2014adam} is utilized to optimize these models and the learning rate is fixed to $10^{-3}$. We train each model for 20 epoches and select the one with the best performance on the development set. Note that we freeze the embedding layer when training all models on SST2, while fine-tune the embedding parameters when training on Yelp. Empirical results show that this can lead to better prediction performance for models on both datasets. We evaluate the test accuracy with the best performing model on the development set, and the performance statistics on SST2 and Yelp dataset are reported on Tab.~\ref{fig:predictionStat}.

\subsection{Baseline Methods}

We evaluate the proposed REAT method by comparing it with five baseline approaches.

\begin{itemize}[leftmargin=*]
\item \noindent\textbf{Vanilla gradient (VanillaGrad)}~\cite{hechtlinger2016interpretation}: Compute gradients of the output prediction with respect to individual entries in word embedding vectors, and use the L2 norm to reduce each vector of the gradients to a single attribution value, representing the contribution of each single word.

\item \noindent\textbf{Integrated gradient (InteGrad)}~\cite{sundararajan2017axiomatic}: Integrate all Vanilla gradients using a linear interpolation between a baseline input and the original input. Here the baseline input are sentences whose word embedding values are all set to zeros.

\item \noindent\textbf{Gradient times input (GradInput)}~\cite{denil2014extraction}: First calculate the gradient of the output with respect to word embedding, and then use dot product of the gradient vector and word embedding vector as the contribution score for a word.

\item \noindent\textbf{LIME}~\cite{ribeiro2016should}: A model-agnostic interpretation method. It approximates the behavior of a RNN in the neighborhood of a given input using an interpretable white-box model. Here, the interpretable model is a sparse linear model.

\item \noindent\textbf{NaiveREAT}: A simplified variant of the proposed method, as introduced in Sec.~\ref{sec:naiveMethod}. The attribution score for a single word is calculated using Eq.~(\ref{equ:baseline}).
\end{itemize}

It is worth noting that except NaiveREAT, the other four baseline methods could only derive word-level contribution scores. To get phrase-level or sentence-level contribution scores, we sum up the scores of all word within a phrase or a sentence.

\begin{figure*}
  \centering
  \includegraphics[width=0.92\linewidth]{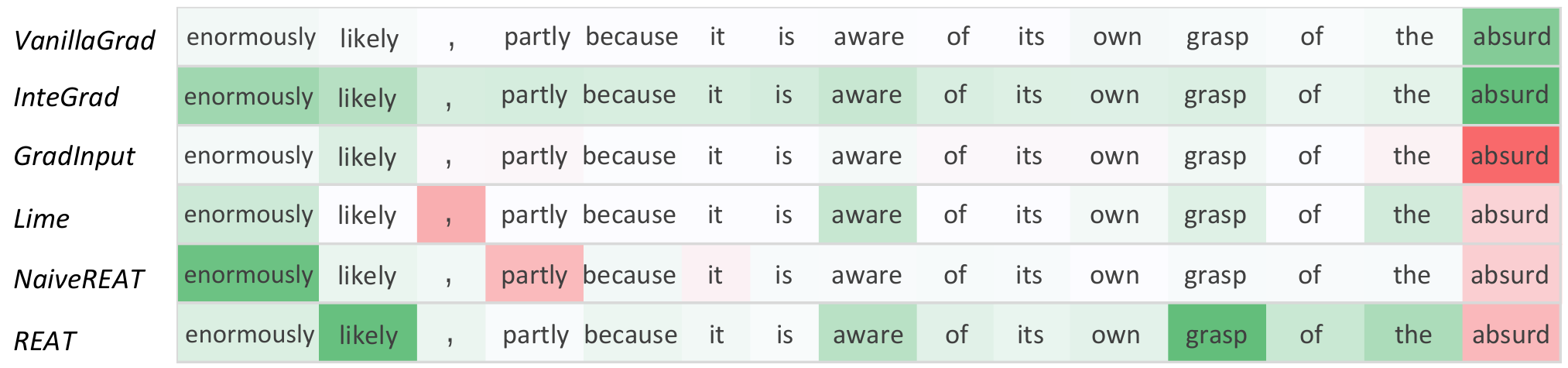}
  \caption{Word-level attribution heatmaps comparing with baseline methods, for a GRU prediction with 99.2\% confidence as positive sentiment. Green and red color denote positive and negative contribution of a word to the prediction, respectively. }
  \label{fig:heatmap1}
\end{figure*}

\subsection{Attribution Faithfulness Evaluation}
In this section, we evaluate the faithfulness of the attribution methods with respect to the target RNN models.
We want to assess whether the attribution results correctly reflect the prediction behavior of RNNs.
In general, the faithfulness of an attribution method is evaluated by deleting the sentence of the highest contribution score and observing the prediction changes of the target RNNs~\cite{nguyen2018comparing}. Specifically, the attribution method first produces contribution scores for sentences in the text. Then, it is expected that once the most important sentence is deleted, it will cause the probability value to significantly drop for the target class. Here we define faithfulness score as the metric:
\begin{equation}
 S_{\text{faithfulness}}  = \frac{1}{N} \sum_{i=1}^N (y(\textbf{x}^{(i)}) - y(\textbf{x}_{\setminus A}^{(i)})),
\label{equ:faithfulness}
\end{equation}
where $A$ denotes the sentence identified as the most predictive for a prediction, and $N$ is the total number of texts in the dataset.
An advantage of this metric is that no knowledge is required of ground truth labels. Theoretically this metric can also be utilized to evaluate word-level and phrase-level attributions. However, empirically, we find some irrelevant words or phrases could also lead to big probability drop, because they cause grammar or syntactic errors instead of really changing the semantics~\cite{miyato2016adversarial,sato2018interpretable}. Therefore, we only use this metric in sentence-level attribution scenarios. 

\begin{table}
\centering
\begin{tabular}{L{1.5cm}C{1.2cm} C{1.2cm} C{1.2cm}}
\toprule 
\textbf{Models} & GRU & LSTM & BiGRU  \\ 
\midrule 
VanillaGrad & 0.272 & 0.243 & 0.068  \\ 
InteGrad & 0.255 & 0.253 & 0.113  \\ 
GradInput & 0.301 & 0.199 & 0.178 \\ 
Lime & 0.209 & 0.188 & 0.092  \\ 
NaiveREAT & 0.213 & 0.207 & 0.114  \\ 
REAT & \textbf{0.311} & \textbf{0.318} & \textbf{0.196}  \\
\bottomrule 
\end{tabular}
\caption{Comparison about attribution faithfulness between our method and the baseline methods.} 
\vspace{-20pt}
\label{tab:faithfulnessEvaluation}
\end{table}

We search in SST2 dataset for texts that contain two sentences and use the word ``but'' as conjunction, and obtain a set with 142 texts.
The faithfulness scores for different attribution methods on SST2 dataset are reported in Tab.~\ref{tab:faithfulnessEvaluation}. The proposed method consistently outperforms the baseline methods for all the three architectures. This result demonstrates two advantages of REAT: (1) the generated interpretations are highly faithful to original RNN, (2) the generated phrase-level interpretations are accurate. Since the NaiveREAT method does not consider the forgetting mechanism, it may assign false positive contribution scores for the first sentence in the testing texts. As a result, it is not faithful to the target model and achieves relatively low faithfulness score comparing to REAT. Besides, for the other baseline methods, they can only output word-level attribution scores. Since the word-level scores are not sufficiently faithful to the target model, the sentence-level scores calculated by summing up the word-level scores will further deviate from the prediction of the target RNN model. As a result, these methods have only limited faithfulness performance. 

\begin{table}
\centering
\begin{tabular}{l c c c c c c}
\toprule 
& \multicolumn{3}{c}{\textbf{SST2}} & \multicolumn{3}{c}{\textbf{Yelp}} \\
\cmidrule(l){2-4} \cmidrule(l){5-7}
\textbf{Models} & GRU & LSTM & BiGRU  & GRU & LSTM & BiGRU\\ 
\midrule 
VanillaGrad & 41.0 & 42.3 & 26.9  & 57.6 & 44.4 & 38.4\\ 
InteGrad & 44.9 & 52.6 & 42.3  & 58.6 & 42.4 & 33.3\\ 
GradInput & 84.6 & 80.8 & 85.9 & 84.8 & 82.8 & 90.9\\ 
Lime & 73.1 & 80.8 & 73.1  & 74.7 & 77.8 & 80.8\\ 
NaiveREAT & 79.5 & 87.2 & \textbf{92.3}  & 55.6 & 83.8 & 86.9\\ 
REAT & \textbf{85.9} & \textbf{89.7} & 88.5  & \textbf{87.9} & \textbf{83.8} & \textbf{92.9}\\
\bottomrule 
\end{tabular}
\caption{Interpretability statistical comparison (in percent) of our method with baseline RNN attribution methods.}
\label{tab:interpretabilityEvaluation}
\end{table}
\subsection{Attribution Interpretability Evaluation}
We evaluate the interpretability of the proposed method to show whether the generated interpretations are reasonable from some fundamental perspectives of human comprehension. We only search for texts that contain both positive adjective words and negative adjective words with obvious sentiment bias. The searching criterion is whether a word belongs to the human annotated lists containing both positive words\footnote{https://gist.github.com/mkulakowski2/4289437} and negative words\footnote{https://gist.github.com/mkulakowski2/4289441}~\cite{liu2005opinion}. We obtain a testing set with 78 and 99 samples from SST2 and Yelp respectively. Then we generate a attribution score after feeding each text sample to a RNN, and evaluate the consistency of attribution with human annotations. Here, we focus on analyzing the attribution scores of these words for the positive sentiment side for all testing texts. The interpretation is considered to be a match if the attribution scores for positive words are lager than negative words, otherwise it is treated as a mismatch. Take text ``It is ridiculous, of course but it is also refreshing'' for example, if the attribution method assigns a higher contribution score to positive word ``refreshing'' comparing to negative word ``ridiculous'', we consider it as a match. The final interpretability score is judged by the ratio of matched cases:
\begin{equation}
 S_{\text{interpretability}}  =  \frac{\# \text{match}}{\# \text{match}+\# \text{mismatch}}.
\label{equ:Interpretability}
\end{equation}

We compare the interpretability score of the proposed method with baseline methods on three RNN architectures over SST2 and Yelp dataset. The results are presented in Tab.~\ref{tab:interpretabilityEvaluation}. The proposed method ranks highest for five tasks among all six classification tasks. We observe that the interpretability scores for the NaiveREAT method, which is introduced in Sec.~\ref{sec:naiveMethod}, are unstable across different models and datasets, partly due to the reason that it is not faithful to the original recurrent model. Sometimes, NaiveREAT could accurately capture the contribution score for strong sentiment word, such as for BiGRU prediction on SST2. In other cases, it may assign some false positive and false negative attribution values, leading to lower interpretability scores. Another interesting finding is that the interpretability of InteGrad does not outperform VanillaGrad consistently on all cases. Besides, GradInput yields consistently better performance comparing to VanillaGrad.

\begin{figure}
  \centering
  \includegraphics[width=1.0\linewidth]{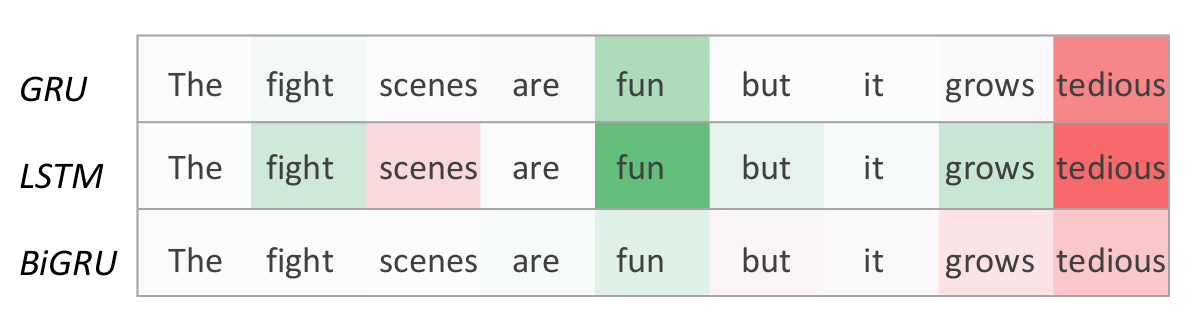}
  \caption{Word-level attribution heatmaps for 3 RNN architectures. Green and red color denote positive and negative contribution of a word to the prediction, respectively. }
  \label{fig:heatmap2}
\end{figure}
\begin{figure*}
  \centering
  \includegraphics[width=0.9\linewidth]{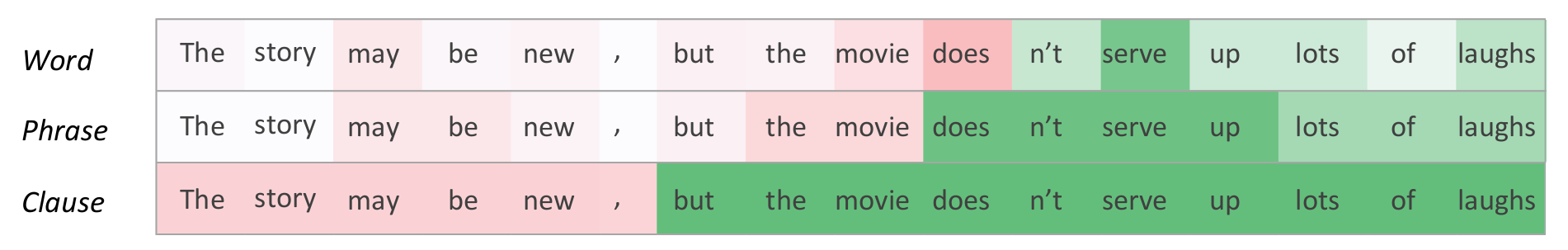}
  \caption{Visualization heatmaps for hierarchical attribution. Green and red color denote positive and negative contribution of a word to the prediction, respectively. }
  \label{fig:heatmap3}
\end{figure*}
\subsection{Qualitative Evaluation via Case Studies}

We provide several case studies to qualitatively check the effectiveness of the proposed method using heatmaps shown from Fig.~\ref{fig:heatmap1} to Fig.~\ref{fig:heatmap3}. We use green color to denote positive contribution and red color for negative contribution. Deeper color means higher contribution to the prediction. 

\subsubsection{Visualization Comparisons with Baseline Methods}
We present attribution visualization for a prediction made by GRU with 99.2\% confidence for positive sentiment, and compare it with the baseline attribution methods. The results are shown in Fig.~\ref{fig:heatmap1}. The heatmap shows that REAT not only identifies that words ``enormously'', ``likely'', ``aware'', ``grasp'' have positive contribution for the prediction, but also captures that ``absurd'' has negative contribution. It is consistent with human comprehension towards this sentence. In contrast, VanillaGrad and InteGrad fail to attribute the negative word; GradInput fails to identify the words that strongly and positively contribute to the prediction, thus can not explain why GRU gives such a high positive prediction score to this text. Also, LIME and NaiveREAT generate some noisy negative scores for irrelevant words, such as ``partly'', ``it'' and ``,''.

\subsubsection{Visualizations Under Different RNN Architectures}
We compare the word-level attribution results of three RNN architectures. For a given text ``The fight scenes are fun but it grows tedious'', GRU, LSTM and BiGRU give positive prediction (51.6\% confidence), positive prediction (96.2\% confidence), and negative prediction (62.7\% confidence), respectively. We display the attribution heatmaps for positive prediction for all three architectures in Fig.~\ref{fig:heatmap2}. GRU gives nearly the same absolute value of attribution score for ``fun'' and ``tedious''. LSTM attributes more words positive contributions than negative words, while BiGRU gives more words negative contribution scores. These attribution heatmaps well reflect the prediction scores, which thus indicates that the interpretations could give users understandable rationale for the predictions.

\subsubsection{Visualization for Hierarchical Attribution}
We also give the visualization heatmaps of hierarchical attributions in Fig.~\ref{fig:heatmap3}. In this case, we show the attribution scores of a negative sentiment prediction with 99.46\% confidence from a LSTM model. For the word-level attribution, ``does'' has a negative contribution for the prediction, while the combination ``does n't serve up'' will have a strong positive contribution for the prediction. The clause-level attribution shows that the first clause has a relatively small negative contribution for prediction, while the second clause has a strong positive contribution for prediction. This hierarchical attention thus could represent the contributions at different levels of granularity.

\subsection{Linguistic Patterns Analysis}
In this section, we apply REAT to analyze the linguistic patterns, in order to comprehend what kind of linguistic knowledge in natural language has been captured by RNNs. We aim to understand whether the linguistic patterns learned are consistent with human cognition.
Overall, the analysis of linguistic patterns is implemented by obtaining the contribution score distributions for different syntactic categories. We first tag each word in a text into different part-of-speech (POS) categories. Here, the tagging is performed using the NLTK package\footnote{https://www.nltk.org/}. We then proceed by feeding that text to the RNN model and use REAT to generate the contribution score for each word. For each text, the class of interest is its prediction label generated by the RNN model.
After that, we summarize the contribution scores for each syntactic category by averaging the scores for all words in that category contained in the dataset.

\begin{figure*}
  \centering
  \begin{overpic}[width=0.9\linewidth]{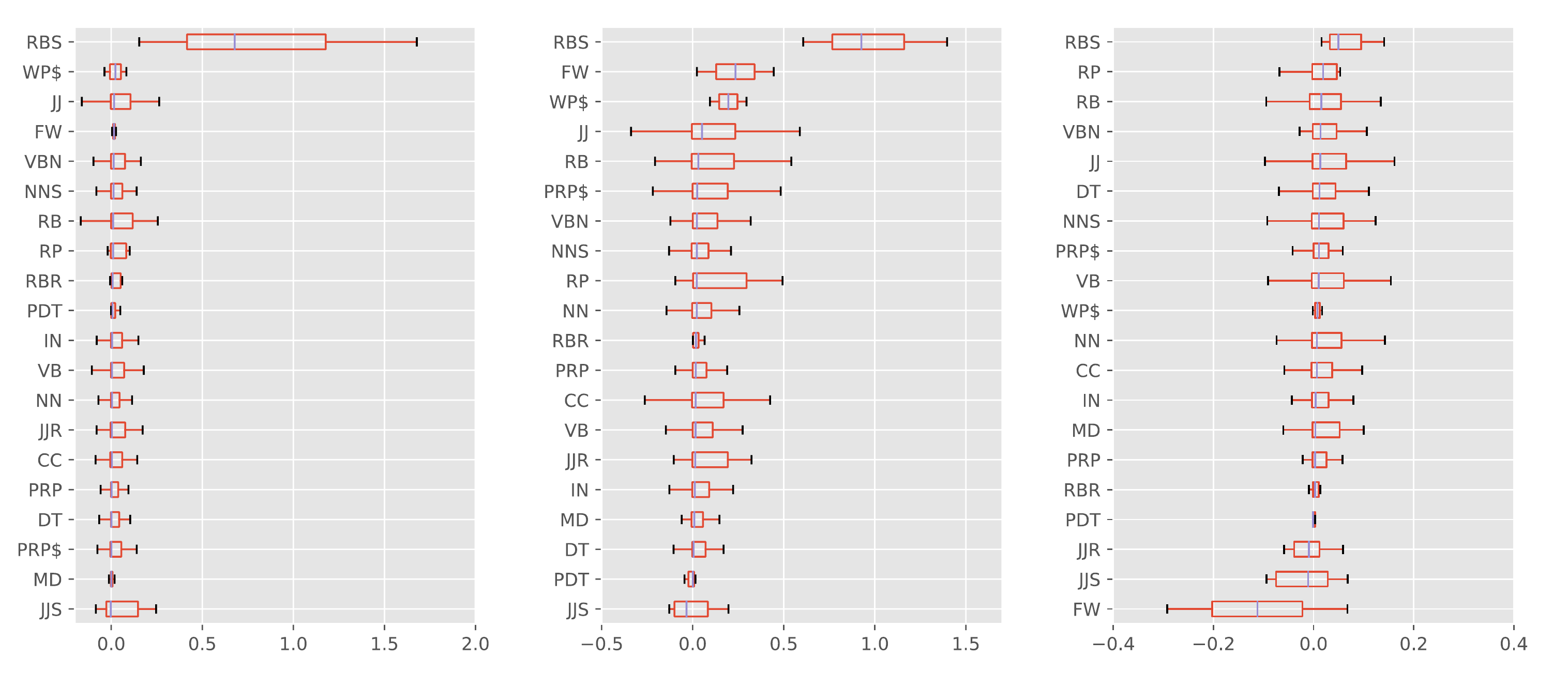}
    \put(15,  -0.5){\footnotesize (a) GRU}
    \put(46,  -0.5){\footnotesize (b) LSTM}
    \put(80,  -0.5){\footnotesize (c) BiGRU}
  \end{overpic}
  \caption{POS category score distributions of three RNN architectures. The x axis denotes attribution scores.}
  \label{fig:linguistic}
\end{figure*}

We employ the SST2 test set to analyze linguistic patterns for the three RNN architectures, and use boxplot to illustrate the distribution patterns, as illustrated in Fig.~\ref{fig:linguistic}. Note that we rank different categories according to the median value. Besides, we combine different verb forms (e.g., past tense and present tense) to a single one, since we observe these verb variants have similar distribution patterns. Also, to capture the primary contribution of each POS category towards the prediction, we omit the outlier points when drawing the boxplot, i.e., those data points that are located outside the fences of the boxplot.
An interesting discovery from the boxplots is that, the RBS category, which denotes superlative adverb such as ``best'' and ``most'', achieves the highest median value on all three RNN architectures. 
This is consistent with our human cognition since we indeed rely on these superlative words to express strong emotions. Besides, the adjective category (JJ) ranks relatively high among all the three architectures. The noun category (NN) has a near-zero median for all three architectures. It indicates that these recurrent models are more sensitive to adjectives than nouns, which also makes sense to common cognition habits. In addition, we can observe that the attribution score distribution patterns are similar between GRU and LSTM, while they are very distinct from that of BiGRU. This is partly due to the reason that BiGRU has a bidirectional structure, which will make it use different evidence in making predictions. These findings demonstrate that these RNNs have learned some useful patterns and pay special attention to words that well represent the underlying task. 

\begin{figure}
  \centering
  \begin{overpic}[width=0.8\linewidth]{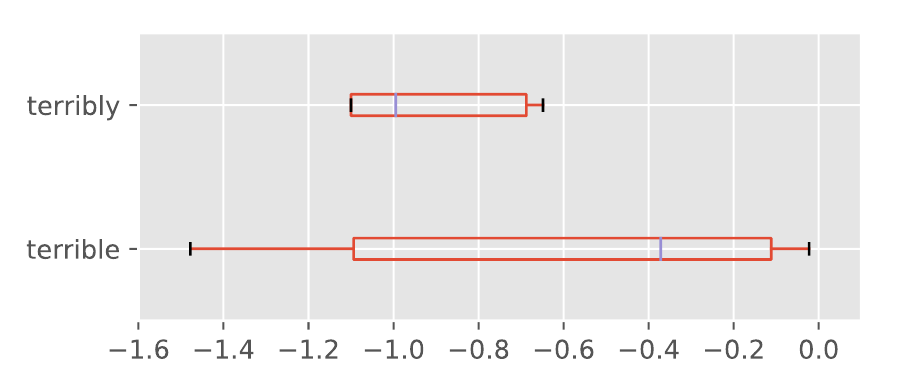}
  \end{overpic}
  \caption{Attribution score distribution for two words. }
  \vspace{-5pt}
  \label{fig:debug}
\end{figure}

\subsection{Model Misbehavior Analysis}
In this section, we apply REAT as a debugging tool to analyze the misbehavior of RNNs. This is motivated by the observation that RNN models may not always meet the expectations of humans, since their performances are still not perfect (see Tab.~\ref{fig:predictionStat}). In this case, the attribution method can be utilized as the tool to analyze their vulnerability and reason the failure cases.

\subsubsection{Attribution-based Failure Case Debugging}
We use a sample text in the SST2 test set to illustrate this process. The LSTM gives a 99.97\% negative sentiment prediction for a text ``Schweiger is talented and terribly charismatic, qualities essential to both movie stars and social anarchists''. However, this prediction makes no sense to humans, since it is in fact a strong positive sentiment text. Using the proposed attribution method, we find that ``terribly'' makes the highest contribution for this negative prediction. One possible explanation is that the LSTM model only captures the meaning similar to ``terrible'', and ignores its other meanings relevant to ``extremely'', since ``terribly'' is a polysemant. To validate this hypothesis, we further generate the attribution score distribution for two words ``terrible'' and ``terribly'' (see Fig.~\ref{fig:debug}), and find that these two words have consistently negative contribution for positive sentiment. It comes as no surprise, since we use pre-trained word2vec as word embeddings, which are fixed and context-free representations and are limited in performance. Besides, the training set only contains a limited number of apperances of ``terribly''. This makes the model fail to learn language polysemy and capture the context dependent meaning of ``terribly''.

\subsubsection{Attribution-based Adversarial Attack} Based on the above attribution analysis, we can perform adversarial attacks using interpretable perturbation. Recent work shows that DNNs are fragile, where a small perturbation could dramatically change their prediction results~\cite{goodfellow2014explaining,papernot2016limitations,liu2018adversarial}. We change the word ``terribly'' to its synonyms ``extremely'' and ``very'', as shown in Tab.~\ref{tab:adversarial}. As the result, the prediction for this text shifts from strong negative to strong positive for both conditions, even though the real meaning of the original text is unchanged. Beside this example, such adversarial attack finding applies to other texts in SST2 dataset. We search for texts in test set of SST2, which contain word ``extremely'' and are originally assigned positive predictions by the LSTM classifier. We change the word ``extremely'' in the original text to ``terribly'', as can be seen in Tab.~\ref{tab:adversarial1}. Although the modified texts are semantically identical to our humans, both predictions of these two samples switch from strong positive to strong negative, demonstrating to some extent the vulnerability of this LSTM classifier. 

\begin{figure}
  \centering
  \begin{overpic}[width=0.95\linewidth]{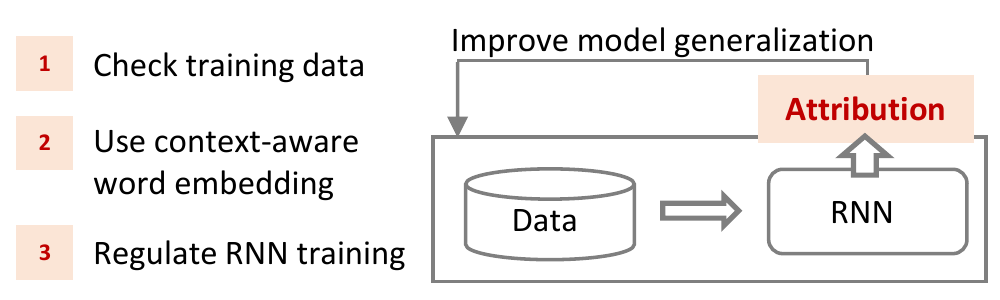}
  \end{overpic}
  \caption{Utilizing attribution to improve models. }
  \vspace{-15pt}
  \label{fig:improveModel}
\end{figure}
\subsubsection{Discussion of Attribution-based Model Generalization Enhancement} The derived attribution method can also be employed to promote the generalization performance of RNN classifiers. Although all the three pretrained RNN classifiers could achieve 100\% accuracy on training set of SST2, all of their accuracy on the test set are below 82\%. This big gap indicates that the RNN classifiers have overfitted to the training set and may have memorized the artifacts and biases that widely exist in texts. Through examining the failure reasons using the the proposed attribution method, or through attacking the RNN classifiers using attribution-based adversarial samples, we could identify the problem of the RNN classifier. We further propose three directions to promote the generalization ability of RNN classifiers, from the perspective of the training data, word embedding or recurrent model, respectively, as illustrated in Fig.~\ref{fig:improveModel}. First, if attribution statistic analysis shows that the RNN has learned lots of bias from the training data, then we can check the training data to reduce the data imbalance or data leaking problem to make sure the training data is less biased. Second, if attribution analysis tell us that trained RNN cannot capture context dependent meanings of words, then we can replace context-free and fixed word embeddings, such as word2vec~\cite{mitkov2005oxford} and GloVe~\cite{pennington2014glove}, to context-aware embeddings, such as the recently proposed ELMo~\cite{peters2018deep}, ULMFiT~\cite{howard2018fine} and BERT~\cite{devlin2018bert}. Third, if the attribution analysis indicates that the RNN heavily relies on superficial correlations in the training data, then we can add a regularizer to regulate the training behavior of the original RNN model. Bridging attribution with improving model generalization is a challenging topic, which is beyond the length of this paper and will be further investigated in our future research.

\begin{table*}
\centering
\begin{tabular}{L{1.3cm}L{12.5cm} L{2.5cm}}
\toprule
&\textbf{Text}&\textbf{Prediction}\\
\midrule
 Original&\small{Schweiger is talented and \colorbox{blue!30}{terribly} charismatic, qualities essential to both movie stars and social anarchists.} & Negative (99.97\%)\\
 Adversarial& \small{Schweiger is talented and \colorbox{blue!30}{extremely} charismatic, qualities essential to both movie stars and social anarchists.} & Positive (81.29\%)\\
 Adversarial&\small{Schweiger is talented and \colorbox{blue!30}{very} charismatic, qualities essential to both movie stars and social anarchists.}&Positive (99.53\%)\\
\bottomrule
\end{tabular}
\caption{Interpretable adversarial attack for a LSTM classifier based on attribution.}
\vspace{-18pt}
\label{tab:adversarial}
\end{table*}

\begin{table*}
\centering
\begin{tabular}{L{1.3cm}L{12.8cm} L{2.5cm}}
\toprule
&\textbf{Text}&\textbf{Prediction}\\
\midrule
 Original&\small{Occasionally melodramatic, it 's also \colorbox{blue!30}{extremely} effective.} & Positive (99.74\%)\\
 Adversarial& \small{Occasionally melodramatic, it 's also \colorbox{blue!30}{terribly} effective.} & Negative (99.00\%)\\
 Original&\small{\colorbox{blue!30}{Extremely} well acted by the four primary actors, this is a seriously intended movie that is not easily forgotten.}&Positive (99.98\%)\\
 Adversarial& \small{\colorbox{blue!30}{Terribly} well acted by the four primary actors, this is a seriously intended movie that is not easily forgotten.} & Negative (87.70\%)\\
\bottomrule
\end{tabular}
\caption{Interpretable adversarial attack for a LSTM classifier based on attribution.}
\vspace{-20pt}
\label{tab:adversarial1}
\end{table*}

\section{Related Work}
Interpretable machine learning is a research topic covering a wide range of directions~\cite{du2018techniques,che2016interpretable,liu2019representation,yang2018towards,karpathy2015visualizing,yuan2019interpreting}.
In this section, we focus on post-hoc and local interpretation methods which are most relevant to ours and briefly review  methods that could provide interpretations for RNN predictions. These attribution methods can be further classified into the following three main categories.

\noindent\textbf{Back-propagation Based Methods}
This line of work calculates the gradient or its variants of the output with respect to the input, for a particular class of interest, using back-propagation to derive the contribution of words. The philosophy is to identify the words whose variation will most significantly lead to the change of output probability. In the simplest case, the gradient signal is back-propagated, which is first proposed in CNN image classification~\cite{simonyan2013deep} and later introduced to provide attribution for text classification in NLP. In NLP, gradients are computed with respect to individual entries in word embedding vectors, and then the L2 norm~\cite{hechtlinger2016interpretation} or the dot product of the gradient and the word embedding~\cite{denil2014extraction} is calculated to reduce the gradient vector to a scalar, representing the contribution of a single word. Besides gradient signal, some work proposes to back-propagate different signals to the input, such as the relevance of the final prediction score through each layer of the network onto the input layer~\cite{bach2015pixel,arras2017explaining}, or only considering the positive gradient signals in the back-propagation process~\cite{springenberg2014striving}.

\noindent\textbf{Perturbation Based Methods}
The motivation of perturbation based methods is that the most important word for a prediction, once perturbed, will cause the largest probability drop of the output for the target class. We can implement the perturbation in two ways: omission~\cite{li2016understanding,kadar2017representation} and occlusion~\cite{poerner2018evaluating}. For omission, the word is deleted directly~\cite{kadar2017representation}. As for occlusion, the word is replaced with a baseline input. Here, a zero-valued word embedding is utilized~\cite{poerner2018evaluating}. However, both omission and occlusion could make the sentence nonsensical. Since word order is an essential factor of RNN predictions, either omission or occlusion can destroy the word order and thus may trigger the adversarial side of RNN. Therefore, it cannot guarantee that it is meaningful interpretations.

\noindent\textbf{Local Approximation Based Methods}
These methods are based on the assumption that behaviors of a complex and opaque model around the neighborhood of a given input can be approximated by a simple and interpretable white-box model~\cite{ribeiro2016should}. A sparse linear model, for example, can be utilized as the locally interpretable model and the weight vector of the linear model is used as the contribution score for a RNN prediction. Sometimes, even the local behavior of a complex model can be extremely non-linear, where linear explanation could lead to poor performance. Therefore, models that are able to capture the non-linear relationship are utilized as the local approximation model. For example, a local attribution method can be designed using if-then rules~\cite{ribeiro2018anchors}. Experimental results demonstrate that the generated rules could be utilized to explain both the current instance and some other relevant instances.

The most similar work to ours are two decomposition based methods~\cite{murdoch2017automatic,murdoch2018beyond}. However, their work are exclusively designed for LSTM, while our attribution method is widely applicable to different recurrent neural network architectures.
\vspace{-7pt}
\section{Conclusion}
We propose a new RNN attribution method, called REAT, to provide interpretations for RNN predictions. REAT decomposes the prediction of a RNN as the additive contribution of each words in the input text, in order to faithfully calculate the response of RNN to the input. REAT could also generate phrase-level attribution scores, which can be combined with syntactic parsing algorithms towards attribution at varying granularity. We apply REAT to three standard RNN architectures, including GRU, LSTM and BiGRU. Empirical results on two sentiment analysis datasets validate that interpretations generated by REAT are both interpretable to humans and faithful to the original RNN classifier.
We further demonstrate that REAT can reveal the useful linguistic patterns learned by RNNs. In addition, we find that RNN models sometimes capture biases in the training data, which causes them to be vulnerable to adversarial attacks. Leveraging the debugging ability provided by REAT, we are able to identify the problems of RNN, and point out several promising directions to improve RNN's generalization ability. This is a challenging topic and will be investigated in our future research. 
\vspace{-10pt}
\begin{acks}
The authors thank the anonymous reviewers for their helpful comments. The work is in part supported by NSF grants IIS-1657196, IIS-1718840 and DARPA grant N66001-17-2-4031. The views and conclusions contained in this paper are those of
the authors and should not be interpreted as representing any funding agencies.
\end{acks}

\bibliographystyle{ACM-Reference-Format}
\bibliography{main}

\end{document}